\documentclass[10pt,twocolumn,letterpaper]{article}

\usepackage{iccv}
\usepackage{times}
\usepackage{epsfig}
\usepackage{graphicx}
\usepackage{amsmath}
\usepackage{amssymb}

\usepackage{algorithm}
\usepackage{algorithmic}
\usepackage{booktabs} 
\usepackage{bm}
\usepackage{mathtools}
\usepackage{makecell}
\usepackage{enumitem}
\usepackage{caption}
\usepackage{subcaption}
\usepackage{lipsum}
\usepackage{cuted}
\usepackage{siunitx}

\usepackage[pagebackref=true,breaklinks=true,letterpaper=true,colorlinks,bookmarks=false]{hyperref}

\iccvfinalcopy 

\def\iccvPaperID{5264} 

\ificcvfinal\fi
\begin{document}

\title{ACNet: Strengthening the Kernel Skeletons for Powerful CNN via Asymmetric Convolution Blocks}

\author{Xiaohan Ding \textsuperscript{1} \quad Yuchen Guo \textsuperscript{2} \quad Guiguang Ding \textsuperscript{1} \quad Jungong Han \textsuperscript{3} \\
	\textsuperscript{1} Beijing National Research Center for Information Science and Technology (BNRist); \\School of Software, Tsinghua University, Beijing, China \\
	\textsuperscript{2} Department of Automation, Tsinghua University; \\Institute for Brain and Cognitive Sciences, Tsinghua University, Beijing, China \\
	\textsuperscript{3} WMG Data Science, University of Warwick, Coventry, United Kingdom \\
	\tt\small dxh17@mails.tsinghua.edu.cn \quad yuchen.w.guo@gmail.com \\
	\tt\small dinggg@tsinghua.edu.cn \quad jungonghan77@gmail.com
}

\maketitle
\thispagestyle{empty}


\begin{abstract}
As designing appropriate Convolutional Neural Network (CNN) architecture in the context of a given application usually involves heavy human works or numerous GPU hours, the research community is soliciting the architecture-neutral CNN structures, which can be easily plugged into multiple mature architectures to improve the performance on our real-world applications. We propose Asymmetric Convolution Block (ACB), an architecture-neutral structure as a CNN building block, which uses 1D asymmetric convolutions to strengthen the square convolution kernels. For an off-the-shelf architecture, we replace the standard square-kernel convolutional layers with ACBs to construct an Asymmetric Convolutional Network (ACNet), which can be trained to reach a higher level of accuracy. After training, we equivalently convert the ACNet into the same original architecture, thus requiring no extra computations anymore. We have observed that ACNet can improve the performance of various models on CIFAR and ImageNet by a clear margin. Through further experiments, we attribute the effectiveness of ACB to its capability of enhancing the model's robustness to rotational distortions and strengthening the central skeleton parts of square convolution kernels.
\end{abstract}

\section{Introduction}
\begin{figure*}
	\begin{center}
		\includegraphics[width=\linewidth]{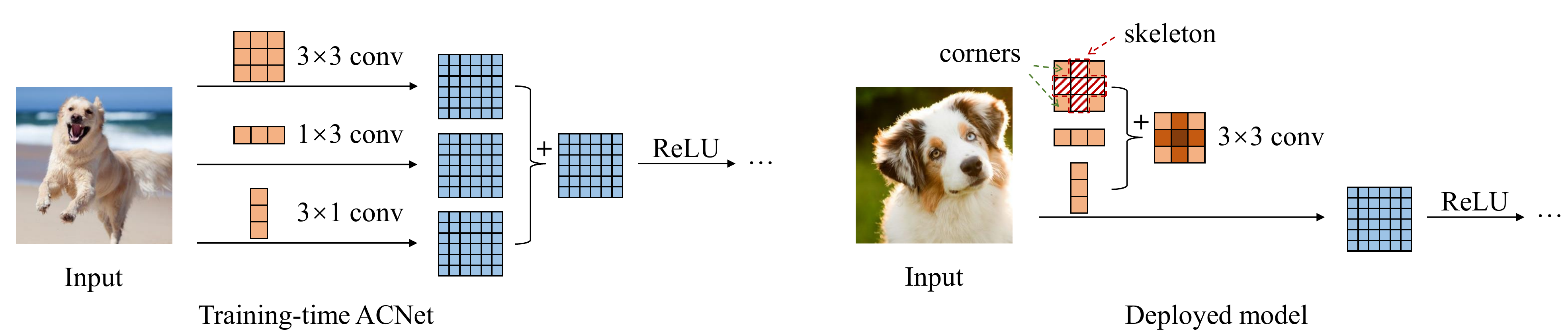}
		\vskip -0.1in
		\caption{Overview of ACNet. For example, we replace every $3\times3$ layer with an ACB comprising three layers with $3\times3$, $1\times3$ and $3\times1$ kernels, respectively, and their outputs are summed up. When the training is completed, we convert the model back into the same structure as the original by adding the asymmetric kernels in each ACB onto the skeleton, which is the crisscross part of the square kernel, as marked on the figure. In practice, this conversion is implemented by building a new model with the original structure and using the converted learned parameters of the ACNet to initialize it.}
		\label{fig-framework}
	\end{center}
\vskip -0.2in
\end{figure*}

Convolutional Neural Network (CNN) has achieved great success in visual understanding, which makes them useful for various applications in wearable devices, security systems, mobile phones, automobiles, \etc. As the front-end devices are usually limited in computational resources and demand real-time inference, these applications require CNN that delivers high accuracy with the constraints of a certain level of computational budgets. Thus it may not be practical to enhance the model by simply employing more trainable parameters and complicated connections. Therefore, we consider it meaningful to improve the performance of CNN with no extra inference-time computations, memory footprint, or energy consumption.

On the other hand, along with the advancements in the CNN architecture designing literature, the performance of the off-the-shelf models has been significantly improved. However, when the existing models cannot meet our specific needs, we may not be allowed to customize a new architecture at the costs of heavy human works or numerous GPU hours \cite{zoph2018learning}. Recently, the research community is soliciting innovative architecture-neutral CNN structures, \eg, SE blocks \cite{hu2018squeeze} and quasi-hexagonal kernels \cite{sun2016design}, which can be directly combined with various up-to-date architectures to improve the performance on our real-world applications.

Some recent investigations on CNN architectures focus on \textbf{1)} how the layers are connected with each other, \eg, simply stacked together \cite{krizhevsky2012imagenet,simonyan2014very}, through identity mapping \cite{he2016deep,szegedy2017inception,zagoruyko2016wide} or densely connected \cite{huang2017densely} and \textbf{2)} how the outputs of different layers are combined to increase the quality of learned representations \cite{ioffe2015batch,szegedy2017inception,szegedy2015going,szegedy2016rethinking}. Considering this, in quest of a generic architecture-neutral CNN structure which can be combined with numerous architectures, we seek to strengthen standard convolutional layers by digging into an orthogonal aspect: \textit{the relationship between the weights and their spatial locations in the kernels}.

In this paper, we propose Asymmetric Convolution Block (ACB), an innovative structure as a building block to replace the standard convolutional layers with square kernels, \eg, $3\times3$ layers, which are widely used in modern CNN. Concretely, for the replacement of a $d\times d$ layer, we construct an ACB comprising three parallel layers with $d\times d$, $1\times d$ and $d \times1$ kernels, respectively, of which the outputs are summed up to enrich the feature space (Fig. \ref{fig-framework}). As the introduced $1\times d$ and $d \times 1$ layers have non-square kernels, we refer to them as the asymmetric convolutional layers, following \cite{szegedy2016rethinking}. Given an off-the-shelf architecture, we construct an Asymmetric Convolutional Network (ACNet) by replacing every square-kernel layer with an ACB and train it until convergence. After that, we equivalently convert the ACNet into the same original architecture by adding the asymmetric kernels in each ACB onto the corresponding positions of the square kernels. Due to the additivity of convolutions with compatible kernel sizes (Fig. \ref{fig-theory}), which is obvious but has long been ignored, the resulting model can produce the same outputs as the training-time ACNet. As will be shown in our experiments (Sect. \ref{sec-improvement-cifar}, \ref{sec-improvement-imagenet}), doing so can improve the performance of several benchmark models on CIFAR \cite{krizhevsky2009learning} and ImageNet \cite{deng2009imagenet} by a clear margin. Better still, ACNet \textbf{1)} introduces NO hyper-parameters, such that it can be combined with different architectures without careful tuning; \textbf{2)} is simple to implement on the mainstream CNN frameworks like PyTorch \cite{paszke2017automatic} and Tensorflow \cite{abadi2016tensorflow}; \textbf{3)} requires NO extra inference-time computational burdens compared to the original architecture.

Through our further experiments, we have partly explained the effectiveness of ACNet. It is observed that a square convolution kernel distributes its learned knowledge unequally, as the weights on the central crisscross positions (which are referred to as the \textit{``skeleton''} of the kernel) are usually larger in magnitude, and removing them causes higher accuracy drop, compared to those in the corners. In each ACB, we add the horizontal and vertical kernels onto the skeletons, thus explicitly making the skeletons more powerful, following the nature of square kernels. Interestingly, the weights on the corresponding positions of the square, horizontal and vertical kernels are randomly initialized and have a possibility to grow opposite in sign, thus summing them up may result in a stronger or weaker skeleton. However, we have empirically observed a consistent phenomenon that the model always learn to enhance the skeletons at every layer. This observation may shed light on future researches on the relationship among the weights at different spatial locations. The codes are available at \url{https://github.com/ShawnDing1994/ACNet}.

Our contributions are summarized as follows.
\begin{itemize}[noitemsep,nolistsep,,topsep=0pt,parsep=0pt,partopsep=0pt]
	\item We propose to use asymmetric convolutions to explicitly enhance the representational power of a standard square-kernel layer in a way that the asymmetric convolutions can be fused into the square kernels with NO extra inference-time computations needed, rather than approximate a square-kernel layer like many prior works \cite{denton2014exploiting,jaderberg2014speeding,jin2014flattened,lo2018efficient,paszke2016enet,szegedy2016rethinking}.
	\item We propose ACB as a novel architecture-neutral CNN building block. We can construct an ACNet by simply replacing every square-kernel convolutional layer in a mature architecture with an ACB without introducing any hyper-parameters, such that its effectiveness can be combined with the numerous advancements in the CNN architecture designing literature. 
	\item We have improved the accuracy of several common benchmark models on CIFAR-10, CIFAR-100, and ImageNet by a clear margin.
	\item We have justified the significance of skeletons in standard square convolution kernels and demonstrated the effectiveness of ACNet in enhancing such skeletons.
	\item We have shown that ACNet can enhance the model's robustness to rotational distortions, which may inspire further studies on the rotational invariance problem.
\end{itemize}

\section{Related work}

\subsection{Asymmetric convolutions}
Asymmetric convolutions are typically used to approximate an existing square-kernel convolutional layer for compression and acceleration. Some prior works \cite{denton2014exploiting,jaderberg2014speeding} have shown that a standard $d\times d$ convolutional layer can be factorized as a sequence of two layers with $d\times 1$ and $1\times d$ kernels to reduce the parameters and required computations. The theory behind is quite simple: if a 2D kernel has a rank of one, the operation can be equivalently transformed into a series of 1D convolutions. However, as the learned kernels in deep networks have distributed eigenvalues, their intrinsic rank is higher than one in practice, thus applying the transformation directly to the kernels results in significant information loss \cite{jin2014flattened}. Denton \etal \cite{denton2014exploiting} tackled this problem by finding a low-rank approximation in an SVD-based manner then finetuning the upper layers to restore the performance. Jaderberg \etal \cite{jaderberg2014speeding} succeeded in learning the horizontal and vertical kernels by minimizing the $\ell$-2 reconstruction error. Jin \etal \cite{jin2014flattened} applied structural constraints to make the 2D kernels separable and obtained comparable performance as conventional CNN with $2\times$ speed-up.

On the other hand, asymmetric convolutions are also widely employed as an architectural design element to save the parameters and computations. For example, in Inception-v3 \cite{szegedy2016rethinking}, the $7\times7$ convolutions are replaced by a sequence of $1\times7$ and $7\times1$ convolutions. However, the authors found out that such replacement is not equivalent as it did not work well on the low-level layers. ENet \cite{paszke2016enet} also adopted this approach for the design of an efficient semantic segmentation network, where the $5\times5$ convolutions are decomposed, allowing to increase the receptive field with reasonable computational budgets. EDANet \cite{lo2018efficient} used a similar method to decompose the $3\times3$ convolutions, resulting in a 33\% saving in the number of parameters and required computations with minor performance degradation.

In contrast, we use 1D asymmetric convolutions not to factorize any layers as part of the architectural designs but enrich the feature space during training and then fuse their learned knowledge into the square-kernel layers.

\subsection{Architecture-neutral CNN structures}

We intend not to modify the CNN architecture but use some architecture-neutral structures to enhance the off-the-shelf models. Thus the effectiveness of our method is supplementary to the advancements achieved by the innovative architectures. Specifically, a CNN structure can be called architecture-neutral if it \textbf{1)} makes no assumptions on the specific architecture, thus can be applied on various models, and \textbf{2)} brings universal benefits. For example, SE blocks \cite{hu2018squeeze} can be appended after a convolutional layer to rescale the feature map channels with learned weights, resulting in a clear accuracy improvement at reasonable costs of extra parameters and computational burdens. As another example, auxiliary classifier \cite{szegedy2015going} can be inserted into the model to assist in supervising the learning process, which can indeed improve the performance by an observable margin but requires extra human works to tune the hyper-parameters.

By contrast, ACNet introduces \textit{NO hyper-parameters} during training and requires \textit{NO extra parameters or computations} during inference. Therefore, in real-world applications, the developer can use ACNet to enhance a variety of models without exhausting parameter tunings, and the end-users can enjoy the performance improvement without slowing down the inference. Better still, since we introduce no custom structures into the deployed model, it can be future compressed via techniques including connection pruning \cite{guo2016dynamic,han2015learning}, channel pruning \cite{ding2019centripetal,ding2019approximated,liu2017learning,luo2017thinet}, quantization \cite{courbariaux2016binarized,gupta2015deep,rastegari2016xnor}, feature map compacting \cite{wang2017beyond}, \etc.

\section{Asymmetric Convolutional Network}

\subsection{Formulation}

For a convolutional layer with a kernel size of $H\times W$ and $D$ filters which takes a $C$-channel feature map as input, we use $\bm{F}\in\mathbb{R}^{H\times W\times C}$ to denote the 3D convolution kernel of a filter, $\bm{M}\in\mathbb{R}^{U\times V\times C}$ for the input, which is a feature map with a spatial resolution of $U\times V$ and $C$ channels, and $\bm{O}\in\mathbb{R}^{R\times T\times D}$ for the output with $D$ channels, respectively. For the $j$-th filter at such a layer, the corresponding output feature map channel is
\begin{equation}\label{eq-def-conv}
\bm{O}_{:,:,j}=\sum_{k=1}^{C}\bm{M}_{:,:,k}\ast\bm{F}^{(j)}_{:,:,k} \,,
\end{equation}
where $\ast$ is the 2D convolution operator, $\bm{M}_{:,:,k}$ is the $k$-th channel of $\bm{M}$ in the form of a $U\times V$ matrix, and $\bm{F}^{(j)}_{:,:,k}$ is the $k$-th input channel of $\bm{F}^{(j)}$, \ie, a 2D kernel of $H\times W$.

In modern CNN architectures, batch normalizations \cite{ioffe2015batch} are widely adopted to reduce overfitting and accelerate the training process. As a common practice, a batch normalization layer is usually followed by a linear scaling transformation to enhance the representational power. Compared to Eq. \ref{eq-def-conv}, the output channel then becomes
\begin{equation}\label{eq-conv-with-bn}
\bm{O}_{:,:,j}=(\sum_{k=1}^{C}\bm{M}_{:,:,k}\ast\bm{F}^{(j)}_{:,:,k} - \mu_j)\frac{\gamma_j}{\sigma_j} + \beta_j \,,
\end{equation}
where $\mu_j$ and $\sigma_j$ are the values of channel-wise mean and standard deviation of batch normalization, $\gamma_j$ and $\beta_j$ are the learned scaling factor and bias term, respectively.

\subsection{Exploiting the additivity of convolution}

\begin{figure}
	\begin{center}
		\includegraphics[width=0.7\columnwidth]{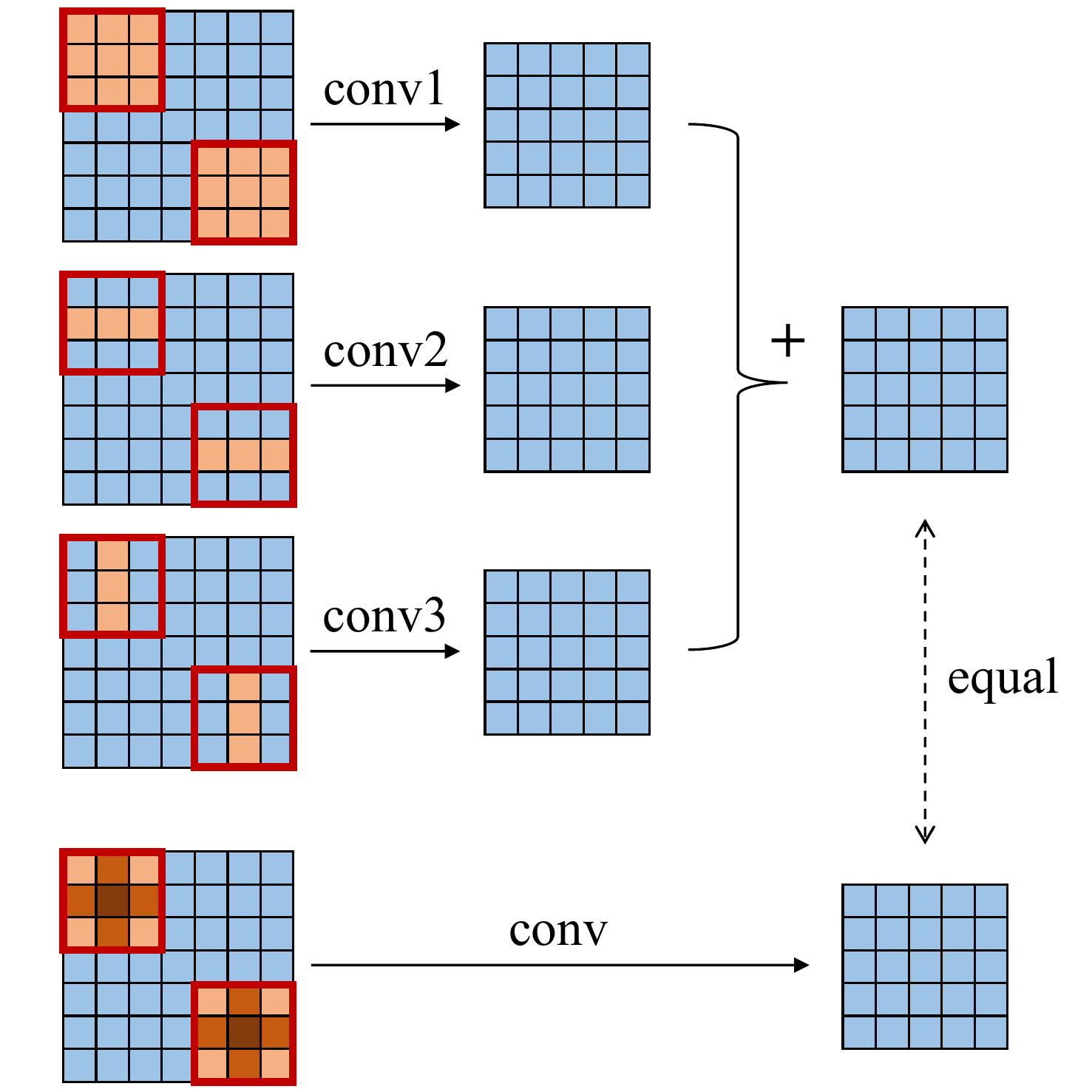}
		\caption{We use sliding windows to provide some intuitions of the additivity of 2D convolutions with different kernel sizes. Here we have three convolutional layers with a kernel size of $3\times3$, $1\times3$ and $3\times1$, respectively, which take the same input. We only depict the sliding window at the top-left and bottom-right corners, for example. It can be observed that the key for the additivity to hold is that the three layers can \textit{share the same sliding window}. Therefore, if we add the kernels of conv2 and conv3 to conv1 on the corresponding positions, using the resulting kernel to operate on the original input will produce the same result, which can be easily verified only using the distributive property of multiplication (Eq. \ref{eq-mult-property}). Best viewed in color.}
		\label{fig-theory}
	\end{center}
\vskip -0.1in
\end{figure}

We seek to employ asymmetric convolutions in a way that they can be equivalently fused into the standard square-kernel layers, such that no extra inference-time computational burdens are introduced. We notice a useful property of convolution: if several 2D kernels with \textit{compatible} sizes operate on the same input with the same stride to produce outputs of the same resolution, and their outputs are summed up, we can add up these kernels \textit{on the corresponding positions} to obtain an equivalent kernel which will produce the same output. That is, the \textit{additivity} may hold for 2D convolutions, even with different kernel sizes,
\begin{equation}\label{eq-additivity}
\bm{I} \ast \bm{K}^{(1)} + \bm{I} \ast \bm{K}^{(2)} = \bm{I} \ast (\bm{K}^{(1)} \oplus \bm{K}^{(2)}) \,,
\end{equation}
where $\bm{I}$ is a matrix, $\bm{K}^{(1)}$ and $\bm{K}^{(2)}$ are two 2D kernels with compatible sizes, and $\oplus$ is the element-wise addition of the kernel parameters on the corresponding positions. Note is that $\bm{I}$ may need to be appropriately clipped or padded.

Here \textit{compatible} means that we can ``patch'' the smaller kernel onto the bigger. Formally, this kind of transformation on layer $p$ and $q$ is feasible if
\begin{equation}
\bm{M}^{(p)}=\bm{M}^{(q)} \,, H_p \leq H_q \,, W_p \leq W_q \,, D_p=D_q \,. 
\end{equation}
\Eg, $3\times1$ and $1\times3$ kernels are compatible with $3\times3$. 

This can be easily verified by investigating the calculation of convolution in the form of sliding windows (Fig. \ref{fig-theory}). For a certain filter with kernel $\bm{F}^{(j)}$, a certain point $y$ on the output channel $\bm{O}_{:,:,j}$ is given by 
\begin{equation}\label{eq-mult-property}
y = \sum_{c=1}^{C}\sum_{h=1}^{H}\sum_{w=1}^{W} F^{(j)}_{h,w,c} X_{h,w,c} \,,
\end{equation}
where $\bm{X}$ is the corresponding sliding window on input $\bm{M}$. Obviously, when we sum up two output channels produced by two filters, the additivity (Eq. \ref{eq-additivity}) holds if for each point $y$ on one channel, its corresponding point on the other channel \textit{shares the same sliding window} $\bm{X}$.

\subsection{ACB for free inference-time improvements} \label{sec-fusion}

\begin{figure}
	\begin{center}
		\includegraphics[width=0.9\columnwidth]{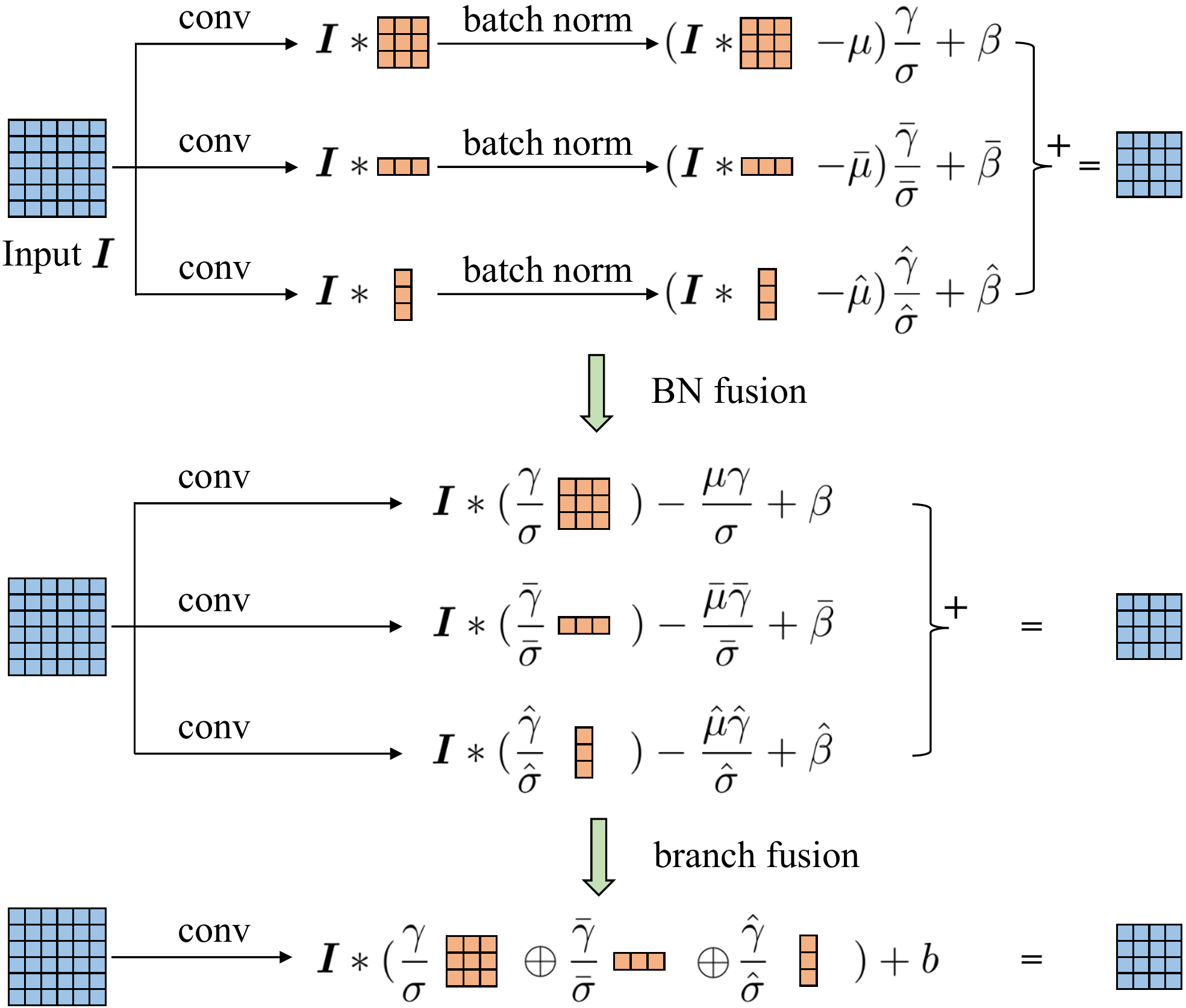}
		\caption{BN and branch fusion. Let $\bm{I}$ be an arbitrary channel of the input feature map $\bm{M}$, for each branch, we first equivalently fuse the parameters of batch normalization into the convolution kernel and a bias term, then add up the fused kernels and bias terms to obtain a single layer.}
		\label{fig-fusion}
	\end{center}
\vskip -0.1in
\end{figure}

In this paper, we focus on $3\times3$ convolutions, which are heavily used in modern CNN architectures. Given an architecture, we construct an ACNet by simply replacing every $3\times3$ layer (together with the following batch normalization layer, if any) with an ACB which comprises three parallel layers with kernel size $3\times3$, $1\times3$ and $3\times1$, respectively. Similar to the common practice in standard CNN, each of the three layers is followed by batch normalization, which is referred to as a branch, and the outputs of three branches are summed up as the output of ACB. Note that we can train the ACNet using the same configurations as the original model without any extra hyper-parameters to be tuned.

As will be shown in Sect. \ref{sec-improvement-cifar} and Sect. \ref{sec-improvement-imagenet}, the ACNet can be trained to reach a higher level of accuracy. When the training is completed, we seek to convert every ACB to a standard convolutional layer which produces identical outputs. By doing so, we can obtain a more powerful network which requires no extra computations, compared to a normally trained counterpart. This conversion is achieved through two steps, namely, BN fusion and branch fusion. 

\paragraph{BN fusion.} The \textit{homogeneity} of convolution allows the following batch normalization and linear scaling transformation to be equivalently fused into the convolutional layer with an added bias. It can be observed from Eq. \ref{eq-conv-with-bn} that for each branch, if we construct a new kernel as $\frac{\gamma_j}{\sigma_j}\bm{F}^{(j)}$ along with an added bias term $-\frac{\mu_j \gamma_j}{\sigma_j} + \beta_j$, we will produce the same output, which can be easily verified.

\paragraph{Branch fusion.} We merge the three BN-fused branches into a standard convolutional layer by adding the asymmetric kernels onto the corresponding positions of the square kernel. In practice, this transformation is implemented by building a network of the original structure and using the fused weights for initialization, thus we can produce the same outputs as the ACNet with the same computational budgets as the original architecture. Formally, for every filter $j$, let $\bm{F}^{\prime(j)}$ be the fused 3D kernel, $b_j$ be the obtained bias term, $\bar{\bm{F}}^{(j)}$ and $\hat{\bm{F}}^{(j)}$ be the kernels of the corresponding filter at the $1\times3$ and $3\times1$ layer, respectively, we have
\begin{equation}\label{eq-fused-kernel}
\bm{F}^{\prime(j)} = \frac{\gamma_j}{\sigma_j}\bm{F}^{(j)} \oplus \frac{\bar{\gamma}_j}{\bar{\sigma}_j}\bar{\bm{F}}^{(j)} \oplus \frac{\hat{\gamma}_j}{\hat{\sigma}_j}\hat{\bm{F}}^{(j)} \,,
\end{equation}
\begin{equation}\label{eq-fused-bias}
b_j = - \frac{\mu_j\gamma_j}{\sigma_j} - \frac{\bar{\mu}_j\bar{\gamma}_j}{\bar{\sigma}_j} - \frac{\hat{\mu}_j\hat{\gamma}_j}{\hat{\sigma}_j} + \beta_j + \bar{\beta}_j + \hat{\beta}_j \,.
\end{equation}

Then we can easily verify that for an arbitrary filter $j$,
\begin{equation}
\bm{O}_{:,:,j} + \bar{\bm{O}}_{:,:,j} + \hat{\bm{O}}_{:,:,j}=\sum_{k=1}^{C}\bm{M}_{:,:,k}\ast\bm{F}^{\prime(j)}_{:,:,k} + b_j \,,
\end{equation}
where $\bm{O}_{:,:,j}$, $\bar{\bm{O}}_{:,:,j}$ and $\hat{\bm{O}}_{:,:,j}$ are the outputs of the original $3\times3$, $1\times3$ and $3\times1$ branch, respectively. Fig. \ref{fig-fusion} shows an example on a single input channel for more intuitions.

Of note is that though an ACB can be equivalently transformed into a standard layer, the equivalence only holds at inference-time because the training dynamics are different, thus giving rise to different performance. The non-equivalence of the training process is due to the random initialization of kernel weights, and the gradients derived by different computation flows they participate in.

\section{Experiments}
We have conducted abundant experiments to verify the effectiveness of ACNet in improving the performance of CNN across a range of datasets and architectures. Concretely, we pick an off-the-shelf architecture as the baseline, build an ACNet counterpart, train it from scratch, convert it into the same structure as the baseline, and test it to collect the accuracy. For the comparability, all the models are trained until the complete convergence, and every pair of baseline and ACNet uses identical configurations, \eg, learning rate schedules and batch sizes.

\subsection{Performance improvements on CIFAR}\label{sec-improvement-cifar}

In order to preliminarily evaluate our method on various CNN architectures, we experiment with several representative benchmark models including Cifar-quick \cite{snoek2012practical}, VGG-16 \cite{simonyan2014very}, ResNet-56 \cite{he2016deep}, WRN-16-8 \cite{zagoruyko2016wide} and DenseNet-40 \cite{huang2017densely} on CIFAR-10 and CIFAR-100 \cite{krizhevsky2009learning}.

For Cifar-quick, VGG-16, ResNet-56, and DenseNet-40, we train the models using a staircase learning rate of 0.1, 0.01, 0.001 and 0.0001 following the common practice. For WRN-16-8, we follow the training configurations reported in the original paper \cite{zagoruyko2016wide}. We use the data augmentation techniques adopted by \cite{he2016deep}, \ie, padding to $40\times40$, random cropping and left-right flipping. 

As can be observed from Table. \ref{table-exp-cifar10} and Table. \ref{table-exp-ch},
the performance of all the models is consistently lifted by a clear margin, suggesting that the benefits of ACBs can be combined with various architectures.

\begin{table}
	\caption{Top-1 accuracy of ACNets and the normally trained baselines on CIFAR-10.}
	\label{table-exp-cifar10}
	\vspace{-0.15in}
	\begin{center}
		\begin{small}
			\begin{tabular}{lccccccc}
				\toprule
				Model		&	Base Top-1					&ACNet Top-1		&Top-1 $\uparrow$\\
				\midrule 
				Cifar-quick	&	83.13						&	84.24		&	1.11	\\
				VGG			&	94.12						&	94.47		&	0.35	\\
				ResNet-56	&	94.31						&	95.09		&	0.78	\\
				WRN-16-8	&	95.56						&	96.15		&	0.59	\\
				DenseNet-40	&	94.29						&	94.84		&	0.55	\\
				\bottomrule
			\end{tabular}
		\end{small}
	\end{center}
\vskip -0.1in
\end{table}
\begin{table}
	\caption{Top-1 accuracy of ACNets and the normally trained baselines on CIFAR-100.}
	\label{table-exp-ch}
	\vspace{-0.15in}
	\begin{center}
		\begin{small}
			\begin{tabular}{lccccccc}
				\toprule
				Model		&	Base Top-1					&ACNet Top-1		&Top-1 $\uparrow$\\
				\midrule
				Cifar-quick	&	53.22						&	54.30		&	1.08\\
				VGG			&	74.56						&	75.20		&	0.64\\
				ResNet-56	&	73.58						&	74.04		&	0.46\\
				WRN-16-8	&	78.65						&	79.44		&	0.79\\
				DenseNet-40	&	73.14						&	73.41		&	0.27\\
				\bottomrule
			\end{tabular}
		\end{small}
	\end{center}
	\vskip -0.1in
\end{table}

\subsection{Performance improvements on ImageNet}\label{sec-improvement-imagenet}
\begin{table*}
	\caption{Accuracy of the ACNet counterparts of AlexNet, ResNets, DenseNet-121 and the baselines on ImageNet.}
	\label{table-exp-imagenet}
	\vspace{-0.15in}
	\begin{center}
		\begin{small}
			\begin{tabular}{lccccccc}
				\toprule
				Model		&	Base Top-1					&ACNet Top-1		&Top-1 $\uparrow$	&	Base Top-5					&ACNet Top-5		&Top-5 $\uparrow$	\\
				\midrule 
				AlexNet		&	55.92						&	\textbf{57.44}		&	1.52	&	79.53	&	\textbf{80.73}		&	1.20		\\
				ResNet-18	&	70.36						&	\textbf{71.14}		&	0.78	&	89.61	&	\textbf{89.96}		&	0.35	\\
				DenseNet-121&	75.15						&	\textbf{75.82}		&	0.67	&	92.45	&	\textbf{92.77}		&	0.32	\\
				\bottomrule
			\end{tabular}
		\end{small}
	\end{center}
\vskip -0.1in
\end{table*}

We then move on to the effectiveness validation of our method on the real-world applications through a series of experiments on ImageNet \cite{deng2009imagenet} which comprises 1.28M images for training and 50K for validation from 1000 classes. We use AlexNet \cite{krizhevsky2012imagenet}, ResNet-18 \cite{he2016deep} and DenseNet-121 \cite{huang2017densely} as the representatives for the plain-style, residual and densely connected architectures, respectively. Every model is trained with a batch size of 256 for 150 epochs, which is longer than the usually adopted benchmarks (\eg, 90 epochs \cite{he2016deep}), such that the accuracy improvement cannot be simply attributed to the incomplete convergence of the base models. For the data augmentation, we employ the standard pipeline including bounding box distortion, left-right flipping and color shift, as a common practice. Especially, the plain version of AlexNet we use comes from the Tensorflow GitHub \cite{Tensorflow-AlexNet}, which is composed of five stacked convolutional layers and three fully-connected layers with no local response normalizations (LRN) or cross-GPU connections. For the faster convergence, we apply batch normalization \cite{ioffe2015batch} on its every convolutional layer. Of note is that since the first two layers of AlexNet use $11\times11$ and $5\times5$ kernels, respectively, it is possible to extend ACBs to have larger asymmetric kernels. However, we still only use $1\times3$ and $3\times1$ convolutions for these two layers, because such large-scale convolutions are becoming less favored in modern CNN, making large ACBs less useful.

As shown in Table. \ref{table-exp-imagenet}, the single-crop Top-1 accuracy of AlexNet, ResNet-18 and DenseNet-121 is lifted by 1.52\%, 0.78\% and 1.18\%, respectively. In practice, aiming at the same target of accuracy, we can use ACNet to enhance a more efficient model to achieve the target with less inference time, energy consumption, and storage space. On the other hand, with the same constraints on computational budgets or model size, we can use ACNet to improve the accuracy by a clear margin such that the gained performance can be viewed as free benefits, from the viewpoint of end-users.

\subsection{Ablation studies}
\begin{table*}
	\caption{Top-1 accuracy of the ACNets with different design configurations and rotational distortions on ImageNet.}
	\vspace{-0.1in}
	\label{table-exp-ablation}
	\begin{center}
		\begin{small}
			\begin{tabular}{lccccccc}
				\toprule
				Model						&	\makecell{Horizontal\\ kernel}	&	\makecell{Vertical\\ kernel}	&	\makecell{BN \\in branch}	&	Original input	& Rotate \ang{90}		&	Rotate \ang{180}		&	Up-down flip 	\\
				\midrule 
				AlexNet						&				&				&				&	55.92			&	28.18				&	31.41				&	31.62			\\
				AlexNet						&				&	\checkmark	&	\checkmark	&	57.10			&	29.65				&	32.86				&	33.02			\\
				AlexNet						&	\checkmark	&				&	\checkmark	&	57.25			&	29.97				&	33.74				&	33.74			\\
				AlexNet						&	\checkmark	&	\checkmark	&	\checkmark	&	\textbf{57.44}	&	\textbf{30.49}		&	\textbf{33.98}		&	\textbf{33.82}	\\
				AlexNet						&	\checkmark	&	\checkmark	&				&	56.18			&	28.81				&	32.12				&	32.33			\\
				\midrule
				ResNet-18					&				&				&				&	70.36			&	41.00				&	41.95				&	41.86			\\
				ResNet-18					&				&	\checkmark	&	\checkmark	&	70.78			&	41.61				&	42.47				&	42.66			\\
				ResNet-18					&	\checkmark	&				&	\checkmark	&	70.70			&	42.06				&	\textbf{43.22}		&	43.05			\\
				ResNet-18					&	\checkmark	&	\checkmark	&	\checkmark	&	\textbf{71.14}	&	\textbf{42.20}		&	42.89				&	\textbf{43.10}	\\
				ResNet-18					&	\checkmark	&	\checkmark	&				&	70.82			&	41.70				&	42.92				&	42.90			\\
				\bottomrule
			\end{tabular}
		\end{small}
	\end{center}
\vskip -0.1in
\end{table*}
\begin{figure}
	\begin{center}
		\includegraphics[width=1\columnwidth]{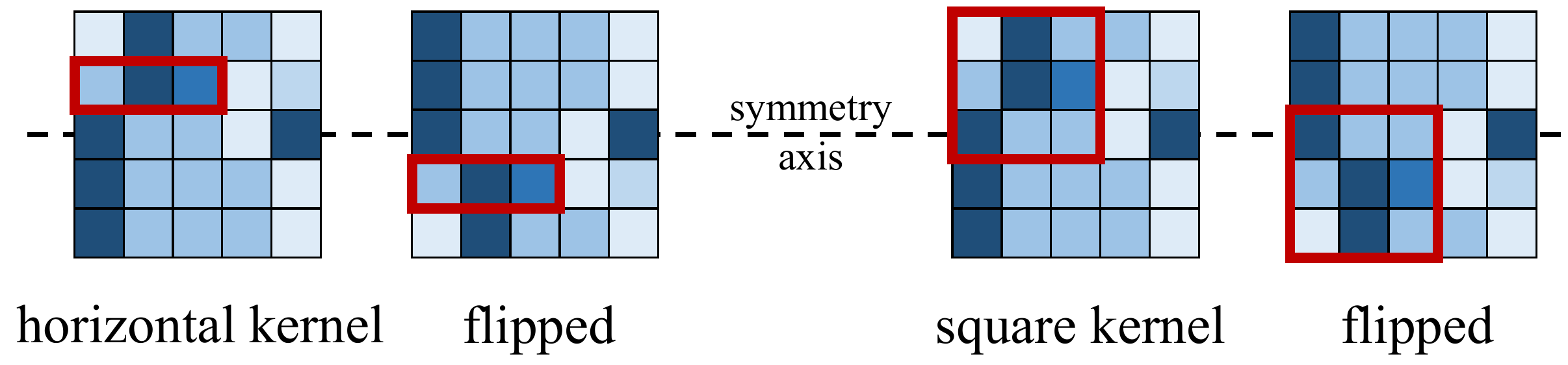}
		\caption{Compared to square kernels, horizontal kernels are more robust to up-down flipping. As shown above, the $1\times3$ kernel will produce the same results on the symmetric positions of the flipped inputs, but the $3\times3$ kernel will not.}
		\label{fig-flipping}
	\end{center}
\vskip -0.2in
\end{figure}

Though we have empirically justified the effectiveness of ACNet, we still desire to find some explanations. In this subsection, we seek to investigate ACNet through a series of ablation studies. Specifically, we focus on the following three design decisions: the usage of \textbf{1)} horizontal kernels, \textbf{2)} vertical kernels, and \textbf{3)} batch normalization in every branch. For the comparability, we train several AlexNet and ResNet-18 models on ImageNet with different ablations using the same training configurations. Of note is that if the batch normalizations in the branches are removed, we batch-normalize the output of the whole ACB instead, \ie, the position of batch normalization layer is changed from pre-summation to post-summation.

As can be observed from Table. \ref{table-exp-ablation}, removing any of the three designs degrades the model. However, though the horizontal and vertical convolutions can both improve the performance, there may exist some difference because the horizontal and vertical directions are treated unequally in practice, \eg, we usually perform random left-right but no up-down image flipping to augment the training data. Therefore, if an upside-down image is fed into the model, the original $3\times3$ layers should produce meaningless results, which is natural, but a horizontal kernel will produce the same outputs as on the original image at the axially symmetric locations (Fig. \ref{fig-flipping}). \Ie, a part of the ACB can still extract the correct features. Considering this, we assume that ACBs may enhance the model's robustness to rotational distortions, enabling the model to generalize better on the unseen data.

We then test the formerly trained models with rotationally distorted images from the whole validation set including counterclockwise \ang{90} rotation, \ang{180} rotation, and up-down flipping. Naturally, the accuracy of every model is significantly reduced, but the models with horizontal kernels deliver observably higher accuracy on the \ang{180} rotated and up-down flipped images. \Eg, the ResNet-18 equipped with only horizontal kernels delivers an accuracy slightly lower than that of the counterpart with only vertical kernels on the original inputs, but 0.75\% higher on the \ang{180} rotated inputs. And when compared with the base model, its accuracy is 0.34\% / 1.27\% higher on the original / \ang{180} flipped images, respectively. Predictably, the models exert similar performance on the \ang{180} rotated and up-down flipped inputs, as \ang{180} rotation plus left-right flipping is equivalent to up-down flipping, and the model is robust to left-right flipping due to the data augmentation methods.

In summary, we have shown that ACBs, especially the horizontal kernels inside, can enhance the model's robustness to rotational distortions by an observable margin. Though this may not be the primary reason for the effectiveness of ACNet, we consider it promising to inspire further researches on the rotational invariance problem.

\subsection{ACB enhances the skeletons of square kernels}

Intuitively, as adding the horizontal and vertical kernels onto the square kernel can be viewed as a means to explicitly enhance the skeleton part, we seek to explain the effectiveness of ACNet by investigating the difference between the skeleton and the weights at the corners.

\begin{figure}[t] 
	\begin{subfigure}{0.90\columnwidth}
		\includegraphics[width=1\linewidth]{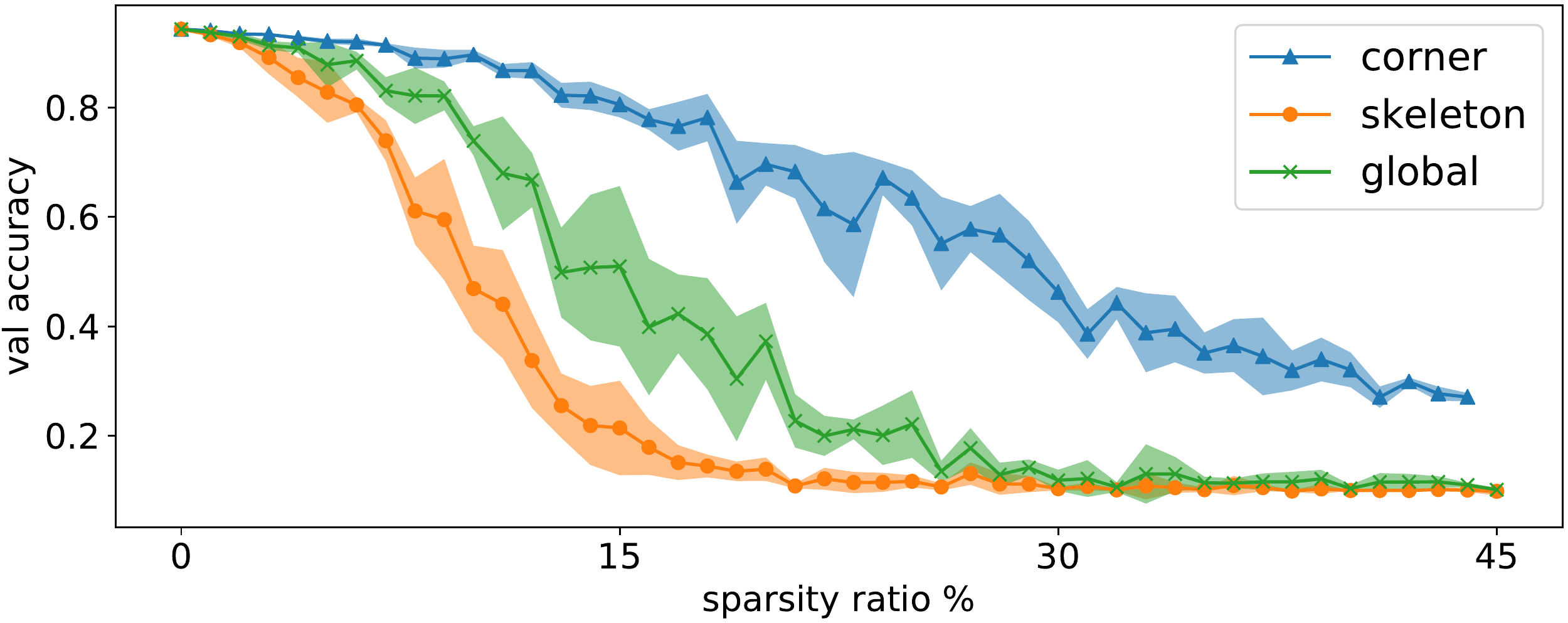} 
		\vskip -0.1in
		\caption{Normally trained ResNet-56.}
		\label{fig-sparsity-normal}
		\vspace{0.1in}
	\end{subfigure}
	\begin{subfigure}{0.90\columnwidth}
		\includegraphics[width=1\linewidth]{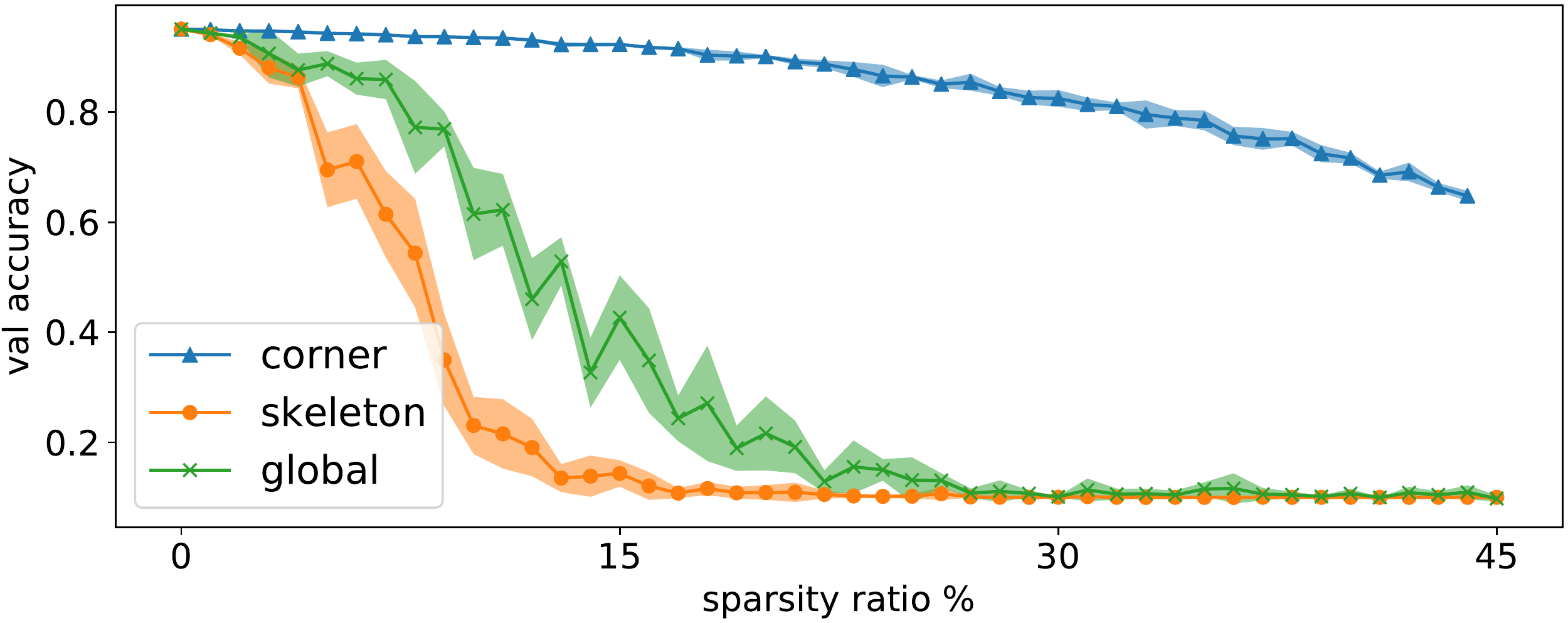} 
		\vskip -0.1in
		\caption{ACNet counterpart of ResNet-56.}
		\label{fig-sparsity-ACB}
		\vspace{0.1in}
	\end{subfigure}
	\begin{subfigure}{0.90\columnwidth}
		\includegraphics[width=1\linewidth]{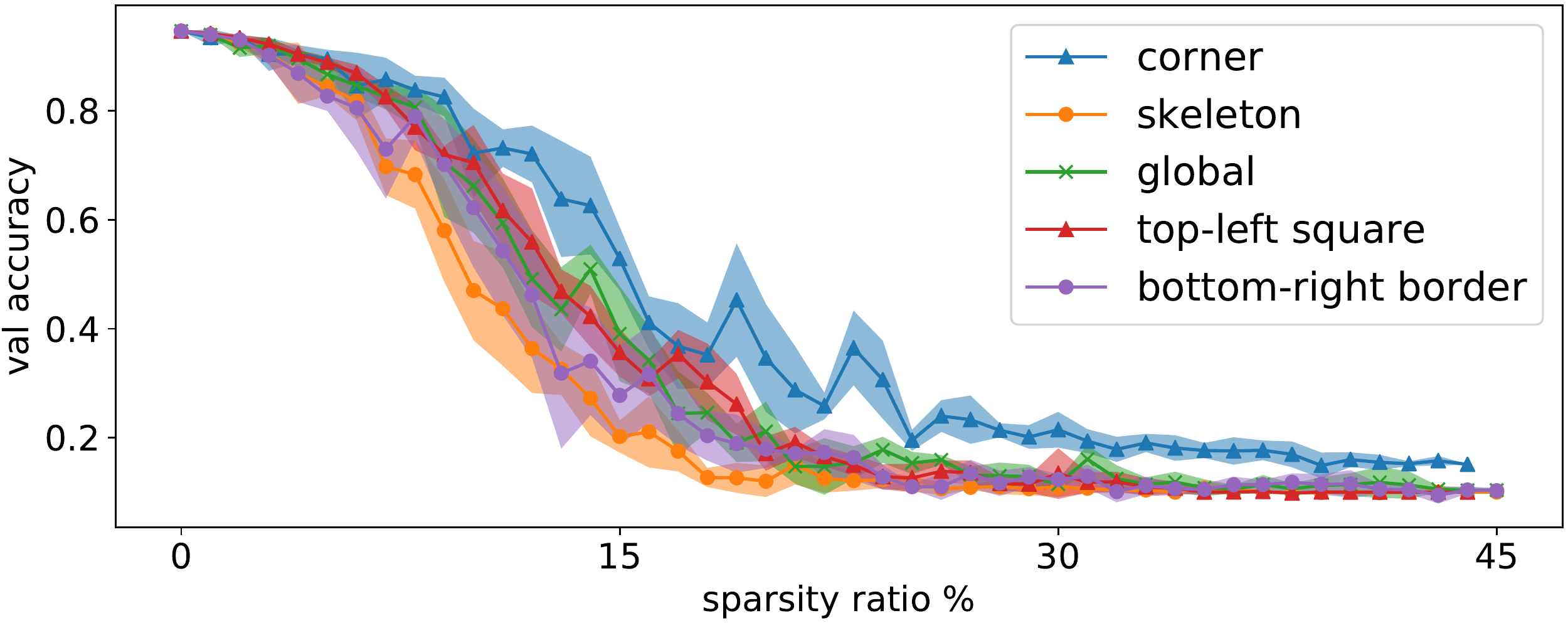} 
		\vskip -0.1in
		\caption{ACNet with the asymmetric kernels added to the border.}
		\label{fig-sparsity-border}
	\end{subfigure}
	\caption{Validation accuracy of different ResNet-56 models on CIFAR-10 with increasing sparsity ratios attained by pruning weights at different locations of the $3\times3$ kernels.}
\vskip -0.1in
\end{figure}

Inspired by the CNN pruning methods \cite{guo2016dynamic,han2015deep,han2015learning}, we start from removing some weights at different spatial locations and observing the performance drop using ResNet-56 on CIFAR-10. Concretely, we randomly set some individual weights in the kernels to zero and test the model. As shown in Fig. \ref{fig-sparsity-normal}, for the curve labeled as \textit{corner}, we randomly select the weights from the four corners of every $3\times3$ kernel and set them to zero in order to attain a given global sparsity ratio of \textit{every} convolutional layer. Note that as $4/9=44.4\%$, a sparsity ratio of 44\% means removing most of the weights at the four corners. For \textit{skeleton}, we randomly select the weights only from the skeleton of every kernel. For \textit{global}, every individual weight in the kernel has an equal chance to be chosen. The experiments are repeated five times with different random seeds, and the \textit{mean}$\pm$\textit{std} curves are depicted.	

As can be observed, all the curves show a tendency of decreasing as the sparsity ratio increases, but not monotonically, due to the random effects. It is obvious that removing the weights from the corners causes less damage to the model, but pruning the skeletons does more harm. This phenomenon suggests that the skeleton weights are more important to the model's representational capacity.

We continue to verify if this observation holds for ACNet. We convert the ACNet counterpart via BN and branch fusion, then conduct the same experiments on it. As shown in Fig. \ref{fig-sparsity-ACB}, we observe an even more significant gap, \eg, pruning almost all the corner weights only degrades the model's accuracy to above 60\%. On the other hand, pruning the skeletons causes more damage, as the model is destroyed when the global sparsity ratio attained by pruning the skeletons merely reaches 13\%, \ie, $13\% \times 9 / 5 = 23.4\%$ weights of the skeletons are removed. 

Then we explore the cause of the above observations by investigating the numeric values of the kernels. We use the magnitude (\ie, absolute value) as the metric for the importance of parameters, which is adopted by many prior CNN pruning works \cite{ding2018auto,guo2016dynamic,han2015learning,li2016pruning}. Specifically, we add up all the fused 2D kernels in a convolutional layer, perform a layer-wise normalization by the max value, and finally obtain an average of the normalized kernels of all the layers. Formally, let $\bm{F}^{(i,j)}$ be the 3D kernel of the $j$-th filter at the $i$-th $3\times3$ layer, $L$ be the number of all such layers, $max$ and $abs$ be the max and element-wise absolute value, respectively, the average kernel magnitude matrix is computed as
\begin{equation}\label{eq-average-kernel}
\bm{A} = \frac{1}{L} \sum_{i=1}^{L} \frac{\bm{S}^{(i)}}{max(\bm{S}^{(i)})} \,,
\end{equation}
where the sum of absolute kernels of layer $i$ is
\begin{equation}
\bm{S}^{(i)} = \sum_{j=1}^{D_i}\sum_{k=1}^{C_i} abs(\bm{F}^{(i,j)}_{:,:,k}) \,.
\end{equation}

\begin{figure}[t] 
	\begin{subfigure}{0.325\columnwidth}
		\includegraphics[page=1,width=0.9\linewidth]{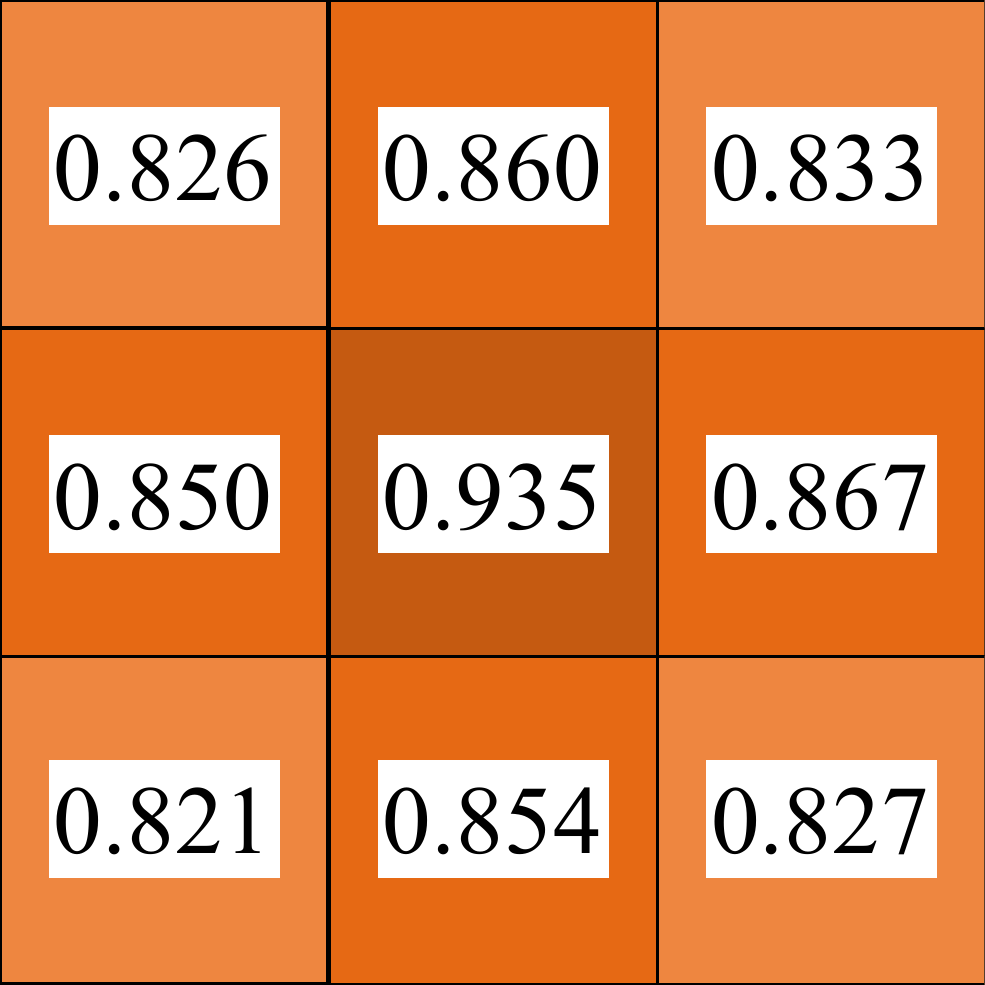} 
		\caption{Normal.}
		\label{fig-accumulate-kernel-normal}
	\end{subfigure}
	\begin{subfigure}{0.325\columnwidth}
		\includegraphics[page=2,width=0.9\linewidth]{rc56-accumulate-kernel.pdf} 
		\caption{ACNet, skeleton.}
		\label{fig-accumulate-kernel-ACB-crisscross}
	\end{subfigure}
	\begin{subfigure}{0.325\columnwidth}
		\includegraphics[page=3,width=0.9\linewidth]{rc56-accumulate-kernel.pdf} 
		\caption{ACNet, border.}
		\label{fig-accumulate-kernel-ACB-border}
	\end{subfigure}
	\caption{The average kernel magnitude matrix $\bm{A}$ of ResNet-56 models trained in different ways on CIFAR-10.}
	\label{fig-accumulat-kernel}
\vskip -0.1in
\end{figure}

We present the $\bm{A}$ values of the normally trained ResNet-56 and the fused ACNet counterpart in Fig. \ref{fig-accumulate-kernel-normal} and Fig. \ref{fig-accumulate-kernel-ACB-crisscross}, where the numeric value and color at a certain grid indicate the average relative importance of the parameter on the corresponding position across all the $3\times3$ layers, \ie, a larger value and darker background color indicates a higher average importance of the parameter.

As can be observed from Fig. \ref{fig-accumulate-kernel-normal}, the normally trained ResNet-56 distributes the magnitude of the parameters in an imbalance manner, \ie, the central point has the largest magnitude, and the points at the four corners have the smallest. Fig. \ref{fig-accumulate-kernel-ACB-crisscross} shows that ACNet aggravates such imbalance, as the $\bm{A}$ values of the four corners are decreased to below 0.400, and the skeleton points have the $\bm{A}$ values above 0.666. In particular, the central point has an $\bm{A}$ value of 1.000, which means that this location has a dominant importance consistently in \textit{every} single $3\times3$ layer. It is noteworthy that the weights on the corresponding positions of the square, horizontal and vertical kernels have a possibility to grow opposite in sign, thus summing them up may result in a larger or smaller magnitude. But we have observed a consistent phenomenon that the model always learn to enhance the skeletons at \textit{every layer}.

We continue to study how the model will behave if we add the asymmetric kernels onto the other positions rather than the central skeletons. Specifically, we train an ACNet counterpart of ResNet-56 using the same training configurations as before, but shift the horizontal convolutions one pixel towards the bottom on the inputs and shift the vertical convolutions towards the right. Accordingly, during branch fusion, we add the BN-fused asymmetric kernels to the bottom-right borders of the square kernels (Fig. \ref{fig-accumulate-kernel-ACB-border}) in order for an equivalent resulting network. It is observed that such ACBs can also enhance the borders, but not as intensively as the regular ACBs do to the skeletons. The model delivers an accuracy of 94.67\%, which is 0.42\% lower than the regular ACNet (Table. \ref{table-exp-cifar10}). Moreover, similar pruning experiments are conducted on the fused model (Fig. \ref{fig-sparsity-border}). As observed, pruning the corners still delivers the best accuracy, and pruning the enhanced bottom-right borders gives no better results than the top-left $2\times2$ squares, \ie, though the magnitudes of the borders have increased, the other parts remain essential to the whole kernels.

In summary: \textbf{1)} the skeletons are inherently more important than the corners in standard square kernels; \textbf{2)} ACB can significantly enhance the skeletons, resulting in improved performance; \textbf{3)} adding the horizontal and vertical kernels to the borders degrades the model's performance compared to regular ACBs; \textbf{4)} doing so can also increase the magnitude of the borders but cannot diminish the importance of the other parts. Therefore, we partly attribute the effectiveness of ACNet to its capability of further strengthening the skeletons. Intuitively, ACNet follows the nature of the square convolution kernels.

\section{Conclusion}

In order to improve the performance of various CNN architectures, we proposed Asymmetric Convolution Block (ACB), which sums up the outputs of three convolutional branches with square, horizontal and vertical kernels, respectively. We construct an Asymmetric Convolutional Network (ACNet) by replacing the square-kernel layers in a mature architecture with ACBs and convert it into the original architecture after training. We have evaluated ACNet by improving various plain-style, residual and densely connected models on CIFAR and ImageNet. We have shown that ACNet can enhance the model's robustness to rotational distortions by an observable margin, and explicitly strengthening the skeletons following the nature of square kernels. Of note is that ACNet introduces NO hyper-parameters to be tuned, requires NO extra inference-time computations, and is simple to implement using mainstream frameworks. 

\section*{Acknowledgement}
This work was supported by the National Key R\&D Program of China (No. 2018YFC0807500), National Natural Science Foundation of China (No. 61571269), National Postdoctoral Program for Innovative Talents (No. BX20180172), and the China Postdoctoral Science Foundation (No. 2018M640131). Corresponding author: Guiguang Ding, Yuchen Guo.

{\small
\bibliographystyle{ieee_fullname}
\bibliography{acgbib}

\begin{thebibliography}{10}\itemsep=-1pt

\bibitem{abadi2016tensorflow}
Mart{\'\i}n Abadi, Paul Barham, Jianmin Chen, Zhifeng Chen, Andy Davis, Jeffrey
  Dean, Matthieu Devin, Sanjay Ghemawat, Geoffrey Irving, Michael Isard, et~al.
\newblock Tensorflow: A system for large-scale machine learning.
\newblock In {\em OSDI}, volume~16, pages 265--283, 2016.

\bibitem{courbariaux2016binarized}
Matthieu Courbariaux, Itay Hubara, Daniel Soudry, Ran El-Yaniv, and Yoshua
  Bengio.
\newblock Binarized neural networks: Training deep neural networks with weights
  and activations constrained to+ 1 or-1.
\newblock {\em arXiv preprint arXiv:1602.02830}, 2016.

\bibitem{deng2009imagenet}
Jia Deng, Wei Dong, Richard Socher, Li-Jia Li, Kai Li, and Li Fei-Fei.
\newblock Imagenet: A large-scale hierarchical image database.
\newblock In {\em Computer Vision and Pattern Recognition, 2009. CVPR 2009.
  IEEE Conference on}, pages 248--255. IEEE, 2009.

\bibitem{denton2014exploiting}
Emily~L Denton, Wojciech Zaremba, Joan Bruna, Yann LeCun, and Rob Fergus.
\newblock Exploiting linear structure within convolutional networks for
  efficient evaluation.
\newblock In {\em Advances in neural information processing systems}, pages
  1269--1277, 2014.

\bibitem{ding2019centripetal}
Xiaohan Ding, Guiguang Ding, Yuchen Guo, and Jungong Han.
\newblock Centripetal sgd for pruning very deep convolutional networks with
  complicated structure.
\newblock In {\em Proceedings of the IEEE Conference on Computer Vision and
  Pattern Recognition}, pages 4943--4953, 2019.

\bibitem{ding2019approximated}
Xiaohan Ding, Guiguang Ding, Yuchen Guo, Jungong Han, and Chenggang Yan.
\newblock Approximated oracle filter pruning for destructive cnn width
  optimization.
\newblock In {\em International Conference on Machine Learning}, pages
  1607--1616, 2019.

\bibitem{ding2018auto}
Xiaohan Ding, Guiguang Ding, Jungong Han, and Sheng Tang.
\newblock Auto-balanced filter pruning for efficient convolutional neural
  networks.
\newblock In {\em Thirty-Second AAAI Conference on Artificial Intelligence},
  2018.

\bibitem{Tensorflow-AlexNet}
GoogLe.
\newblock Tensorflow-alexnet.
\newblock
  \url{https://github.com/tensorflow/models/blob/master/research/slim/nets/alexnet.py},
  2017.

\bibitem{guo2016dynamic}
Yiwen Guo, Anbang Yao, and Yurong Chen.
\newblock Dynamic network surgery for efficient dnns.
\newblock In {\em Advances In Neural Information Processing Systems}, pages
  1379--1387, 2016.

\bibitem{gupta2015deep}
Suyog Gupta, Ankur Agrawal, Kailash Gopalakrishnan, and Pritish Narayanan.
\newblock Deep learning with limited numerical precision.
\newblock In {\em International Conference on Machine Learning}, pages
  1737--1746, 2015.

\bibitem{han2015deep}
Song Han, Huizi Mao, and William~J Dally.
\newblock Deep compression: Compressing deep neural networks with pruning,
  trained quantization and huffman coding.
\newblock {\em arXiv preprint arXiv:1510.00149}, 2015.

\bibitem{han2015learning}
Song Han, Jeff Pool, John Tran, and William Dally.
\newblock Learning both weights and connections for efficient neural network.
\newblock In {\em Advances in Neural Information Processing Systems}, pages
  1135--1143, 2015.

\bibitem{he2016deep}
Kaiming He, Xiangyu Zhang, Shaoqing Ren, and Jian Sun.
\newblock Deep residual learning for image recognition.
\newblock In {\em Proceedings of the IEEE conference on computer vision and
  pattern recognition}, pages 770--778, 2016.

\bibitem{hu2018squeeze}
Jie Hu, Li Shen, and Gang Sun.
\newblock Squeeze-and-excitation networks.
\newblock In {\em Proceedings of the IEEE conference on computer vision and
  pattern recognition}, pages 7132--7141, 2018.

\bibitem{huang2017densely}
Gao Huang, Zhuang Liu, Kilian~Q Weinberger, and Laurens van~der Maaten.
\newblock Densely connected convolutional networks.
\newblock In {\em Proceedings of the IEEE conference on computer vision and
  pattern recognition}, volume~1, page~3, 2017.

\bibitem{ioffe2015batch}
Sergey Ioffe and Christian Szegedy.
\newblock Batch normalization: Accelerating deep network training by reducing
  internal covariate shift.
\newblock In {\em International Conference on Machine Learning}, pages
  448--456, 2015.

\bibitem{jaderberg2014speeding}
Max Jaderberg, Andrea Vedaldi, and Andrew Zisserman.
\newblock Speeding up convolutional neural networks with low rank expansions.
\newblock {\em arXiv preprint arXiv:1405.3866}, 2014.

\bibitem{jin2014flattened}
Jonghoon Jin, Aysegul Dundar, and Eugenio Culurciello.
\newblock Flattened convolutional neural networks for feedforward acceleration.
\newblock {\em arXiv preprint arXiv:1412.5474}, 2014.

\bibitem{krizhevsky2009learning}
Alex Krizhevsky and Geoffrey Hinton.
\newblock Learning multiple layers of features from tiny images.
\newblock 2009.

\bibitem{krizhevsky2012imagenet}
Alex Krizhevsky, Ilya Sutskever, and Geoffrey~E Hinton.
\newblock Imagenet classification with deep convolutional neural networks.
\newblock In {\em Advances in neural information processing systems}, pages
  1097--1105, 2012.

\bibitem{li2016pruning}
Hao Li, Asim Kadav, Igor Durdanovic, Hanan Samet, and Hans~Peter Graf.
\newblock Pruning filters for efficient convnets.
\newblock {\em arXiv preprint arXiv:1608.08710}, 2016.

\bibitem{liu2017learning}
Zhuang Liu, Jianguo Li, Zhiqiang Shen, Gao Huang, Shoumeng Yan, and Changshui
  Zhang.
\newblock Learning efficient convolutional networks through network slimming.
\newblock In {\em 2017 IEEE International Conference on Computer Vision
  (ICCV)}, pages 2755--2763. IEEE, 2017.

\bibitem{lo2018efficient}
Shao-Yuan Lo, Hsueh-Ming Hang, Sheng-Wei Chan, and Jing-Jhih Lin.
\newblock Efficient dense modules of asymmetric convolution for real-time
  semantic segmentation.
\newblock {\em arXiv preprint arXiv:1809.06323}, 2018.

\bibitem{luo2017thinet}
Jian-Hao Luo, Jianxin Wu, and Weiyao Lin.
\newblock Thinet: A filter level pruning method for deep neural network
  compression.
\newblock In {\em Proceedings of the IEEE international conference on computer
  vision}, pages 5058--5066, 2017.

\bibitem{paszke2016enet}
Adam Paszke, Abhishek Chaurasia, Sangpil Kim, and Eugenio Culurciello.
\newblock Enet: A deep neural network architecture for real-time semantic
  segmentation.
\newblock {\em arXiv preprint arXiv:1606.02147}, 2016.

\bibitem{paszke2017automatic}
Adam Paszke, Sam Gross, Soumith Chintala, Gregory Chanan, Edward Yang, Zachary
  DeVito, Zeming Lin, Alban Desmaison, Luca Antiga, and Adam Lerer.
\newblock Automatic differentiation in pytorch.
\newblock In {\em NIPS-W}, 2017.

\bibitem{rastegari2016xnor}
Mohammad Rastegari, Vicente Ordonez, Joseph Redmon, and Ali Farhadi.
\newblock Xnor-net: Imagenet classification using binary convolutional neural
  networks.
\newblock In {\em European Conference on Computer Vision}, pages 525--542.
  Springer, 2016.

\bibitem{simonyan2014very}
Karen Simonyan and Andrew Zisserman.
\newblock Very deep convolutional networks for large-scale image recognition.
\newblock {\em arXiv preprint arXiv:1409.1556}, 2014.

\bibitem{snoek2012practical}
Jasper Snoek, Hugo Larochelle, and Ryan~P Adams.
\newblock Practical bayesian optimization of machine learning algorithms.
\newblock In {\em Advances in neural information processing systems}, pages
  2951--2959, 2012.

\bibitem{sun2016design}
Zhun Sun, Mete Ozay, and Takayuki Okatani.
\newblock Design of kernels in convolutional neural networks for image
  classification.
\newblock In {\em European Conference on Computer Vision}, pages 51--66.
  Springer, 2016.

\bibitem{szegedy2017inception}
Christian Szegedy, Sergey Ioffe, Vincent Vanhoucke, and Alexander~A Alemi.
\newblock Inception-v4, inception-resnet and the impact of residual connections
  on learning.
\newblock In {\em Thirty-First AAAI Conference on Artificial Intelligence},
  2017.

\bibitem{szegedy2015going}
Christian Szegedy, Wei Liu, Yangqing Jia, Pierre Sermanet, Scott Reed, Dragomir
  Anguelov, Dumitru Erhan, Vincent Vanhoucke, and Andrew Rabinovich.
\newblock Going deeper with convolutions.
\newblock In {\em Proceedings of the IEEE conference on computer vision and
  pattern recognition}, pages 1--9, 2015.

\bibitem{szegedy2016rethinking}
Christian Szegedy, Vincent Vanhoucke, Sergey Ioffe, Jon Shlens, and Zbigniew
  Wojna.
\newblock Rethinking the inception architecture for computer vision.
\newblock In {\em Proceedings of the IEEE conference on computer vision and
  pattern recognition}, pages 2818--2826, 2016.

\bibitem{wang2017beyond}
Yunhe Wang, Chang Xu, Chao Xu, and Dacheng Tao.
\newblock Beyond filters: Compact feature map for portable deep model.
\newblock In {\em International Conference on Machine Learning}, pages
  3703--3711, 2017.

\bibitem{zagoruyko2016wide}
Sergey Zagoruyko and Nikos Komodakis.
\newblock Wide residual networks.
\newblock {\em arXiv preprint arXiv:1605.07146}, 2016.

\bibitem{zoph2018learning}
Barret Zoph, Vijay Vasudevan, Jonathon Shlens, and Quoc~V Le.
\newblock Learning transferable architectures for scalable image recognition.
\newblock In {\em Proceedings of the IEEE conference on computer vision and
  pattern recognition}, pages 8697--8710, 2018.

\end{thebibliography}
}

\end{document}